\documentclass{article}
\usepackage{spconf,amsmath,graphicx}
\usepackage{booktabs}

\title{EADNet: Efficient Asymmetric Dilated Network for Semantic Segmentation}

\name{Qihang Yang$^{\star}$ \quad Tao Chen$^{\star}$ \quad Jiayuan Fan$^{\S}$ \quad Ye Lu$^{\star}$ \quad Chongyan Zuo$^{\dagger}$ \quad Qinghua Chi$^{\dagger}$\thanks{Thanks to NSFC (Grant No. 62071127), NSFC (Grant No. U1909207), Shanghai Pujiang Program (No.19PJ1402000) and Huawei Innovation Research Program (HIRP No.HO2019040102003P010) for funding.}}
\address{$^{\star}$ School of Information Science and Technology, Fudan University \\
           $^{\S}$ Academy for Engineering and Technology, Fudan University \\
        $^{\dagger}$ Fusion Platform Development Dept, Shanghai Huawei Technologies Co., Ltd., China}

\begin{document}
\maketitle
\begin{abstract}
Due to real-time image semantic segmentation needs on power constrained edge devices, there has been an increasing desire to design lightweight semantic segmentation neural network, to simultaneously reduce computational cost and increase inference speed. In this paper, we propose an efficient asymmetric dilated semantic segmentation network, named EADNet, which consists of multiple developed asymmetric convolution branches with different dilation rates to capture the variable shapes and scales information of an image. Specially, a multi-scale multi-shape receptive field convolution (MMRFC) block with only a few parameters is designed to capture such information. Experimental results on the Cityscapes dataset demonstrate that our proposed EADNet achieves segmentation mIoU of 67.1\% with smallest number of parameters (only 0.35M) among mainstream lightweight semantic segmentation networks.
\end{abstract}
\section{Introduction}
\label{sec:intro}

\begin{figure}[h]
  \centering
  \includegraphics[width=0.9\linewidth]{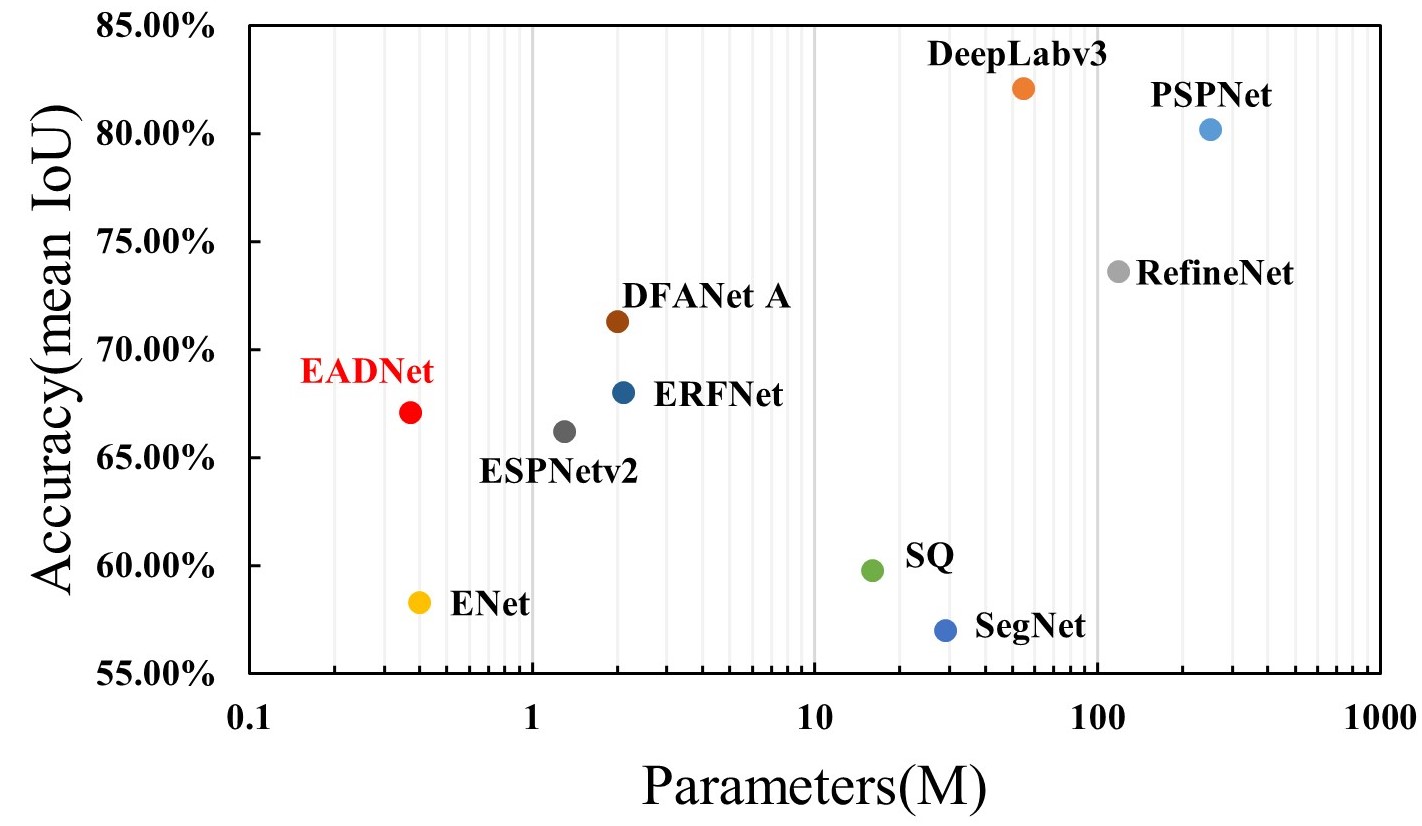}
  \caption{Accuracy vs. Model parameters on Cityscapes.}
  \label{Figure acc vs param}
\end{figure}

Semantic segmentation is a fundamental task in computer vision, and it can be broadly applied to a variety of scenarios. Particularly, such task has strict requirement for model size and response time when deployed in computing power-restrained devices. Figure \ref{Figure acc vs param} shows the parameter sizes and accuracies (mIoUs) for existing state-of-the-art semantic segmentation networks on the Cityscape urban dataset \cite{Cordts2016The}. It can be seen that the high-accuracy semantic segmentation networks \cite{Zhao2017Pyramid, He2016Deep, Lin2017RefineNet} tend to have a larger number of parameters, causing heavier computational cost and often are difficult to meet the real-time segmentation requirement on edge devices. In contrast, some lightweight networks \cite{Romera2018ERFNet, Paszke2016ENet, Mehta2018ESPNet, Mehta2018ESPNetv2} have smaller number of parameters indicating fast inference speed, yet sacrificing the prediction
accuracies. Our proposed EADNet is on the left top of the figure and achieves the best trade-off between accuracy and parameter size amongst peer networks.

Many high precision semantic segmentation networks develop various convolution blocks such as Pyramid Pooling Module in PSPNet \cite{Zhao2017Pyramid}, ASPP in Deeplab series \cite{Chen2017DeepLab}, Object Context Module in OCNet \cite{Yuan2018OCNet} to extract the multi-scale context information to improve the network performance. However, these operations are unsuitable to lightweight networks duo to their massive computation cost.

Lightweight segmentation networks aim to generate high-quality pixel-level segmentation results with limited computational power. ENet \cite{Paszke2016ENet} reduces the times of downsamling and uses depthwise convolution in pursuit of an extremely tight framework. ERFNet \cite{Romera2018ERFNet} uses a non-bottleneck 1D structure to make trade-off between precision and computation and achieves high precision with simple structure. ESPNetv2 \cite{Mehta2018ESPNetv2} introduces EESP Unit to pursuit larger receptive field with less parameters. BiSeNet \cite{Yu2018BiSeNet} proposes spatial path and context path to extract spatial and context information. ICNet \cite{Zhao2017ICNet} uses multi-scale images as input and cascade network to improve efficiency.
\begin{figure}[h]
  \centering
  \includegraphics[width=0.9\linewidth]{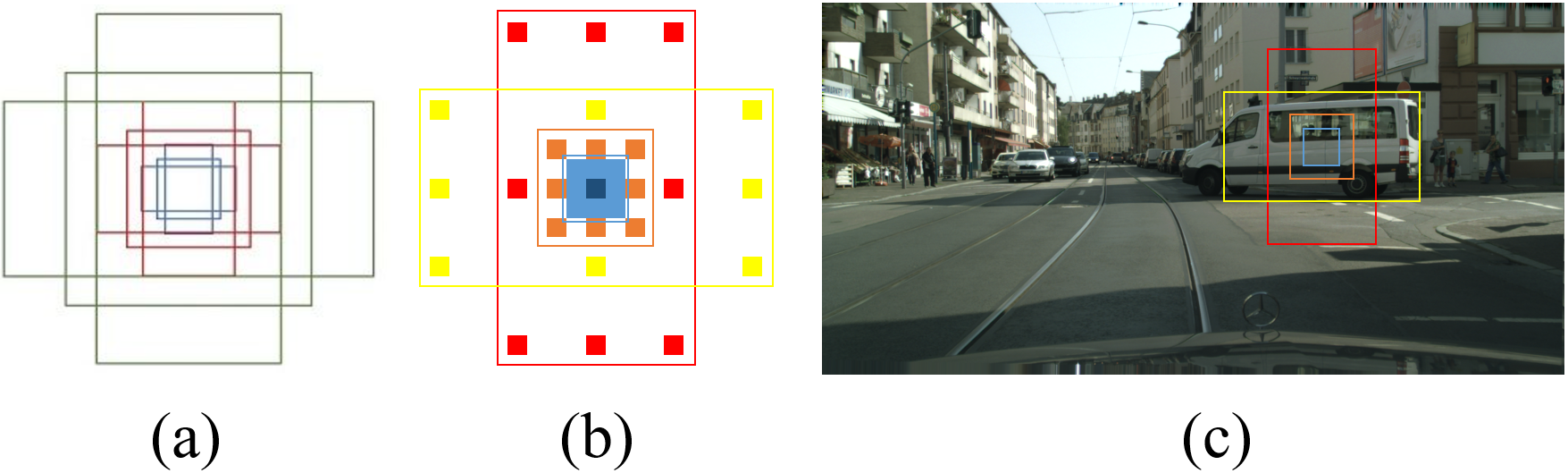}
  \caption{(a) shows the 9 anchors in Faster-RCNN \cite{Ren2015Faster}. (b) shows the four receptive field rectangles of MMRFC block. (c) shows the corresponding image areas for the four receptive field rectangles in (b).}
  \label{Figure receptive field}
\end{figure}

To extract context information with lower computation cost, in this work, we focus on the design of lightweight convolution block to capture multi-scale multi-shape context information and further build a rapid and accurate semantic segmentation network by employing the designed convolution block. Inspired by faster R-CNN \cite{Ren2015Faster},
which utilizes anchor boxes with different aspect ratios to generate region proposals for covering objects of different shapes, we proposed a multi-scale multi-shape receptive field convolution (MMRFC) block, which has receptive fields of different aspect ratios to well match with objects presenting different shapes and scales in the feature maps. Figure \ref{Figure receptive field} shows the 9 anchor boxes used in Faster R-CNN \cite{Ren2015Faster} and the 4 receptive field boxes in our MMRFC block.

 Unlike the lightweight networks \cite{Paszke2016ENet, Romera2018ERFNet, Yu2018BiSeNet, Zhao2017ICNet} which commonly use encoder-decoder structure, by improving the UNet \cite{Ronneberger2015U} structure, we use a specially designed skip connection to combine detailed information in shallow layers and abstract information in deep layers at a small computational cost without decoder stage. The MMRFC block developed as above is incorporated into this structure to form an extreme tight semantic segmentation network, EADNet. EADNet has a very small number of parameters and involves minimum computational operations, and is suitable for deployment on embedded edge devices.

In summary, our contributions are summarized as follows:

1)	We propose multi-scale multi-shape receptive field convolution (MMRFC) block, to capture multi-scale context information which is sensitive to various shapes of objects in input images.

2)	Based on MMRFC block, we develop a lightweight semantic segmentation network, EADNet, which has an extreme tight and efficient structure to combine shallow layer features and deep layer features without decoder stage.

3)	We achieve competitive segmentation accuracy (67.1\% mIoU) with smallest network parameter size (only 0.35M) on the Cityscapes test dataset.

\section{PROPOSED METHOD}

As described before, we propose MMRFC block and further adopt MMRFC block in the specially designed structure to form EADNet. EADNet achieves the optimized point of semantic segmentation in terms of the accuracy, the number of parameters and FLOPs. Following subsections will give detailed analysis of the MMRFC block and EADNet.

\begin{figure}[ht]
  \centering
  \includegraphics[width=0.9\linewidth]{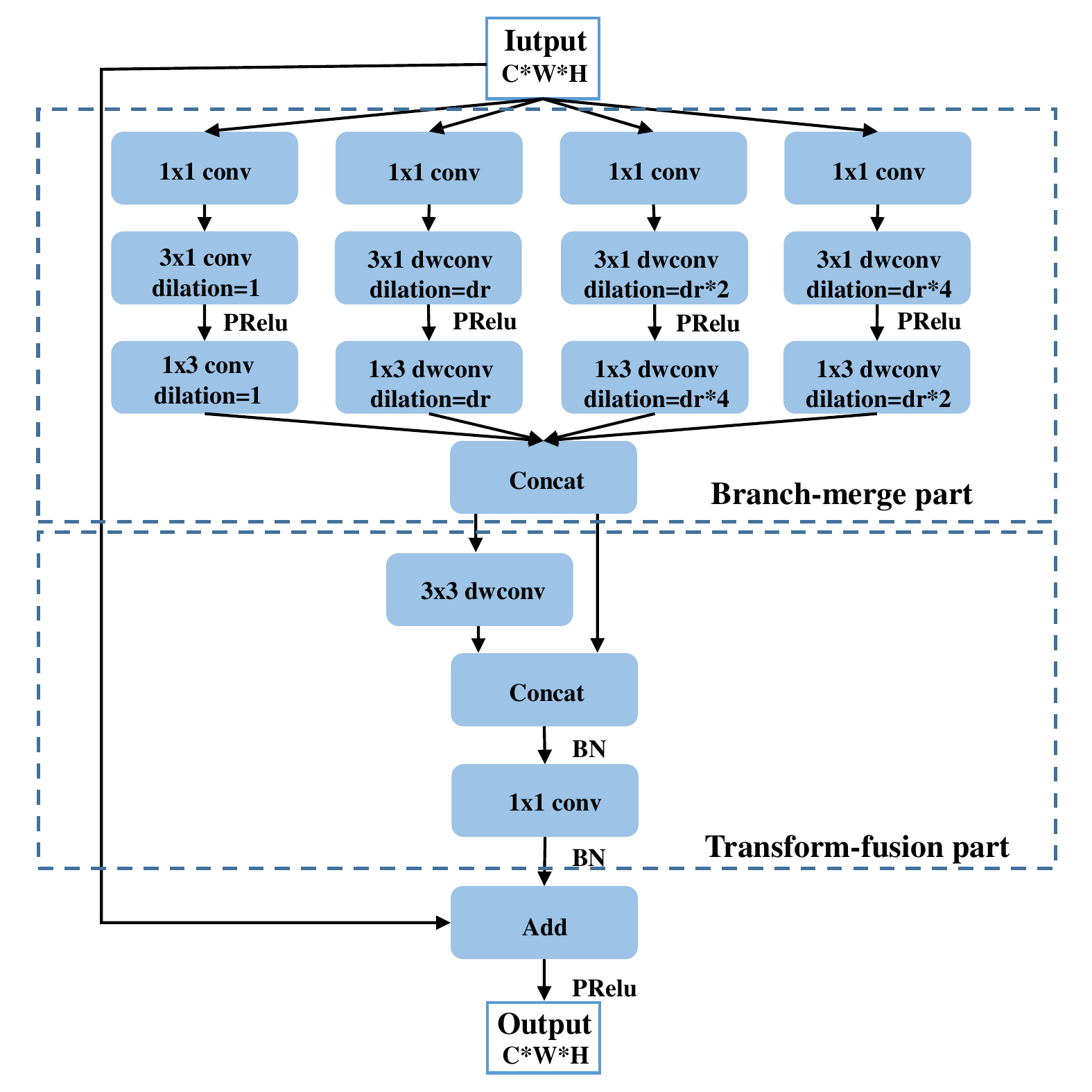}
  \caption{The internal structure of MMRFC block, $dr$ denotes the base dilation rate of current MMRFC block.}
  \label{Figure MMRFC}
\end{figure}

\subsection{MMRFC Block}
MMRFC is designed to capture mulit-scale multi-shape context information with low computation cost and is suitable for lightweight semantic segmentation networks. MMRFC block consists of two parts: branch-merge part and transform-fusion part as shown in Figure \ref{Figure MMRFC}. The branch-merge part contains four convolution branches, where each branch is composed of three sequential convolution operations: one 1$\times$1 pointwise convolution to reduce the input dimension and extract features sensitive to this branch, one 3$\times$1 asymmetric (depthwise) convolution and one 1$\times$3 (depthwise) asymmetric convolution. The two asymmetric convolutions in each branch have same or different dilation rates to achieve variable shapes and sizes of receptive fields, and a parametric rectified linear unit (PRelu) \cite{He2015Delving} is adopted between them. The output of four branches will be concatenated as the input of transform-fusion part.

Transform-fusion part is inspired by GhostNet \cite{han2019ghostnet}, which shows that redundant information is important to improve the accuracy of semantic segmentation and redundant features can be easily generated by linear transform operation. Transform-fusion part is composed of one 3$\times$3 depthwise convolution, a parallel skip connection, a feature concatenation operation and a 1$\times$1 pointwise convolution, to double the channel of input feature maps and fuse their channel information. Lots of new redundant information is generated by the 3$\times$3 depthwise convolution with little additional cost, and finally goes through feature fusion by 1$\times$1 pointwise convolution. 

\begin{figure*}[h]
  \centering
  \includegraphics[width=0.9\textwidth]{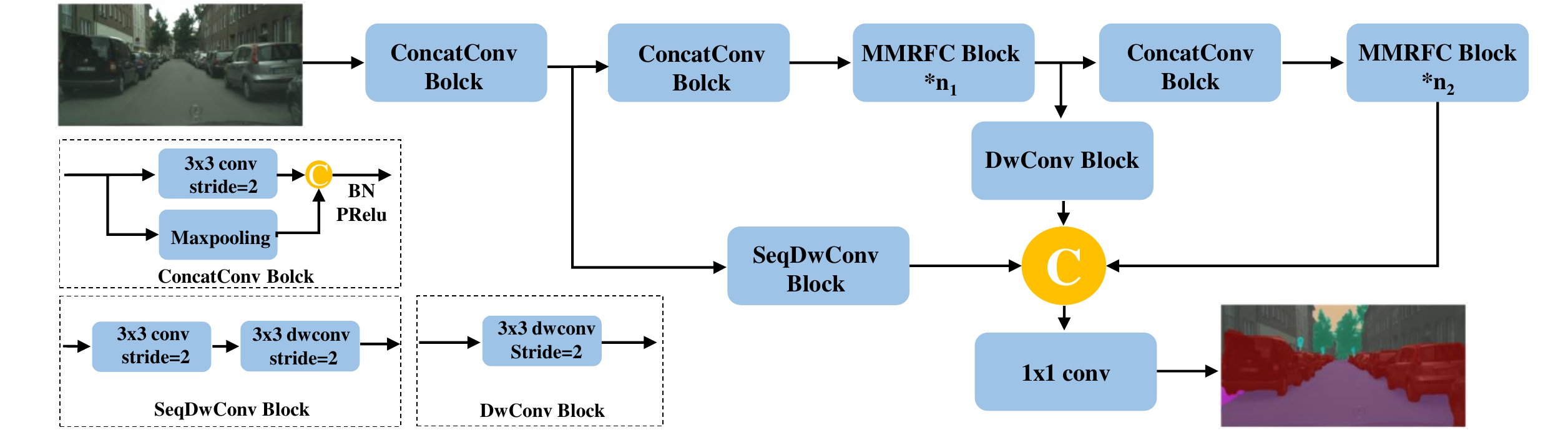}
  \label{Figure EADNet}
  \setlength{\abovedisplayskip}{3pt}
  \setlength{\belowdisplayskip}{3pt}
  \caption{Overview of our EADNet. In figure, ”C” means concatenation, $*n_i$ represents the number of MMRFC blocks, left bottom is the internal structure of ConcatConv blocks, SeqDwConv block and DwConv block.}
\end{figure*}

In asymmetric dilated convolution, when dilation rate gets larger, e.g., beyond 24, the 1$\times$3 or 3$\times$1 filter may degenerate to a 1$\times$1 filter since only the center weight is effective as analyzed in \cite{Chen2017DeepLab}. According to this, we adopt dilation rate in four branches of branch-merge part as (1,1), ($dr$, $dr$), ($dr*2$, $dr*4$) and ($dr*4$, $dr*2$) respectively to generate receptive fields like Figure \ref{Figure receptive field} (b), where $dr$ represents the base dilation rate of the MMRFC block and the value of $dr$ usually increases gradually as the layer goes deeper. To ensure every dilation convolution kernel works, the maximum value of $dr$ is set to 6, yielding a maximum dilation rate of 24 for each asymmetric convolution in all MMRFC blocks. 

 Generally, the spatial and channel correlations between adjacent pixels in an image are stronger than those between non-adjacent pixels. Therefore, we adopt asymmetric convolution to extract spatial and cross-channel features from adjacent pixels in the first branch in branch-merge part. In the second, third and fourth branch which has larger dilation rates, depthwise asymmetric convolutions with less computation cost are used to reduce the computation cost. MMRFC block follows similar strategy as bottleneck structure to first compress the number of input channels to $1/8$ in each of the four branches by a 1$\times$1 convolution, perform concatenated convolution in each branch for feature extraction in branch-merge part, and then combine them and restore the channel number to the original in transform-fusion part. These strategies can significantly reduce the number of FLOPs and parameters in MMRFC block. Following will give the detailed formulas of the FLOPs and parameter size computation.

Assume the shape of input feature is $C*W*H$, where $C$, $W$, $H$ denote the channel number, width and height of the feature, respectively. We denote by $P_{bi} (i=1, 2, 3, 4)$ as the convolution operation parameters in the four branches of branch-merge part. $P_{t}$ and $P_{total}$ are the convolution operation parameters of transform-fusion part and whole MMRFC block. $F_{total}$ denotes the total FLOPs in the convolution operations of MMRFC block. They are computed as follows:

\begin{equation}
P_{bi}\!=\!=\left\{
\begin{array}{r}
\!(C\!+\!1){\times}\frac{C}{8}\!+\!(3{\times}\frac{C}{8}\!+\!1){\times}\frac{C}{8}{\times}2\; for \;i=1\\ \\
\!(C\!+\!1){\times}\frac{C}{8}\!+\!(3\!+\!1){\times}\frac{C}{8}{\times}2 \;for \; i=2, 3, 4
\end{array} \right. 
\end{equation}

\begin{equation}
P_t=(3{\times}3+1){\times}\frac{C}{2}+(C+1){\times}C
\end{equation}

\begin{equation}
  P_{total}=\sum_{i=1,2,3,4}^{i}P_{bi}+P_{t}
\end{equation}

\begin{equation}
F_{total}=P_{total}{\times}W{\times}H
\end{equation}

The maximum number of input channels for MMRFC block is set at 128 in our network. Therefore, in the case of $C\!=\!128$, ignoring a small number of parameters and FLOPs in PRelu \cite{He2015Delving}, batch normalization \cite{ioffe2015batch} and other non-convolution operations, MMRFC block has a parameter size of 27360 and FLOPs of $27360*W*H$, which is $1/5$ of that for the common 3$\times$3 convolution with same input and output dimension.

\begin{table*}
  \caption{ FPS, FLOPs, parameter size and mIoU comparison on Cityscapes test set.}
  \label{Table 2}
  \begin{tabular}{lccccccc}
    \toprule
    Method &Pretrain &Input Size & Inference Time(ms) & FPS & FLOPs(G) &Parameters(M) &mIoU(\%)\\
    \midrule
  
    SegNet \cite{Vijay2017SegNet}& ImageNet& $1024*2048$ & 152.68 & 6.55 & 1310 & 29 & 57.0\\
    SQ \cite{Treml2016Speeding}  & ImageNet & $1024*2048$ & 88.19  & 11.34 & 501  & 16 & 59.8\\
    ERFNet \cite{Romera2018ERFNet}& -       & $1024*2048$ & 43.71  & 22.88 & 103  & 2.1& 68.0\\
    
    ENet \cite{Paszke2016ENet}& -       & $1024*2048$ & 36.9  & 27.11 & 22  & 0.37& 58.3\\
    
    DFANet A\cite{Li_2019_CVPR}   &ImageNet & $1024*2048$ & 26.91  & 37.16 & 28   & 2.0 & \textbf{71.3}\\
    ESPNetv2 \cite{Mehta2018ESPNetv2}& -    & $1024*2048$ & 24.58  & 40.69 & 23.5 & 1.3& 66.2\\
    \textbf{Ours}& - & $1024*2048$ & \textbf{23.98} & \textbf{41.7} & \textbf{18} & \textbf{0.35}& 67.1\\

    \bottomrule
  \end{tabular}
\end{table*}

\subsection{EADNet Structure}

The design of EADNet pursuits extreme lightweight, and its structure is clear and simple as illustrated in Figure 4. We employ three types of downsampling convolution blocks in EADNet: ConcatConv block, SeqDwConv block and DwConv block. It is seen that we insert $n_i$ MMRFC blocks (e.g., denoted by $n_1$, $n_2$ in the figure) after the second and third ConcatConv block, respectively, for efficient feature extraction. The $dr$ of MMRFC blocks are gradually increased as the layer goes deeper to enlarged the receptive field of EADNet and we set $n_1=6$, $n_2=9$ experimentally in our network. 

SeqDwConv block and DwConv block are designed to retain more high-resolution information when downsampling. They make the feature map sizes from shallow layers consistent, which are then concatenated with the final MMRFC block output to compensate the missing of detailed information in deep layers. The concatenated features contain detailed information in shallow layers and semantic information in deep layers and these information is important for network prediction. We use pointwise convolution to predict segmentation map and perform bilinear interpolation to resize the segmentation map to the original resolution of the input image. Compared with similar U-shape structure with computationally complex decoder stage, the EADNet saves a lot of parameters and FLOPs as the decoder is replaced by the pointwise convolution. These strategies mentioned above make our network to have just 0.35M parameters, and 18 GFLOPs when the input resolution is $1024*2048$ pixels.

\begin{table}
  \caption{Experiment result on CamVid test set. }
  \label{Table 3}
  \begin{tabular}{lccc}
    \toprule
     Method &Input Size &FLOPs(G) & mIoU(\%)\\
    \midrule
     SegNet \cite{Vijay2017SegNet}&$960*720$ & 427.34 & 46.4\\
     DFANet A\cite{Li_2019_CVPR} &$960*720$ & 9.03  & 64.7\\
     \textbf{ours}      &$960*720$&  \textbf{5.99}  &\textbf{68.3}\\

  \bottomrule
\end{tabular}
\end{table}

\section{EXPERIMENTS}

\subsection{Experiment Settings}

We evaluated our EADNet in Cityscapes dataset \cite{Cordts2016The} and trained using the Adam \cite{kingma2014adam} optimization of stochastic gradient decent. The “poly” learning rate policy is adopted, meaning the current learning rate is base learning rate multiplied with $(1-\frac{iter}{maxiter})^{0.9}$, where $iter, maxiter$ correspond to current iteration index and the maximum number of iterations respectively. The loss function is the sum of cross-entropy terms for each spatial position in the output score map, and segmentation performances are measured by the common Intersection-over-Union (IoU) metric.

\subsection{Comparison with State-of-the-arts}

\begin{figure}[h]
  \centering
  \includegraphics[width=0.95\linewidth]{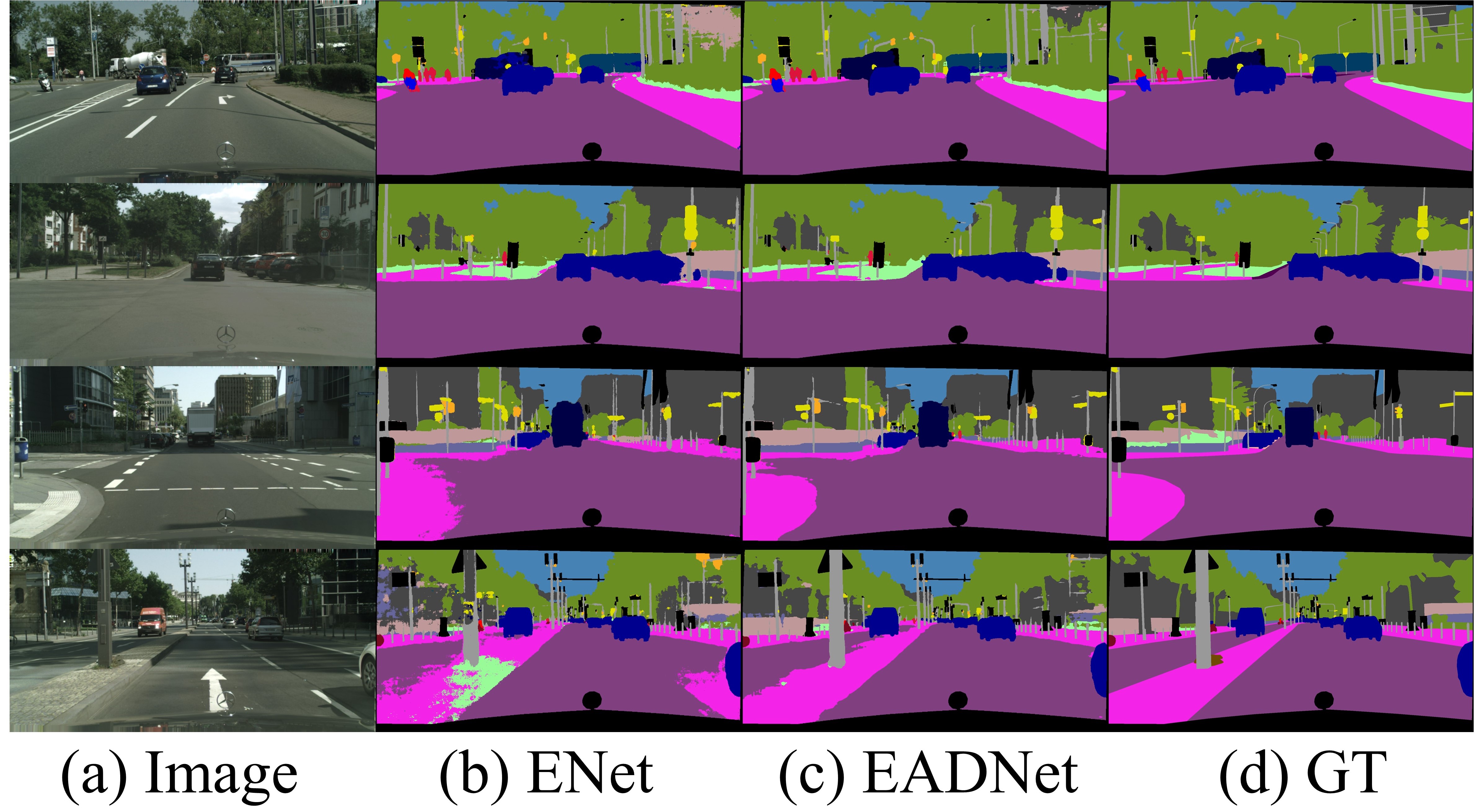}
  \caption{Example results of ENet (b) and EADNet (c). (a) is the images in Cityscapes dataset and (d) is their ground truth. 
}
  \label{Figure compare}

\end{figure}
The comparison results on overall FPS, FLOPs, parameter size and mIoU for state-of-the-art methods on Cityscapes test set are shown in Table \ref{Table 2}. “Pretrain” in the table denotes if the model has been pretrained on external data like ImageNet \cite{Deng2009Imagenet}, FPS is measured on a single Tesla V100. As shown in the table, compared with the existing smallest network ENet \cite{Paszke2016ENet} which has 0.37M parameters and 22 GFLOPs, our model has fewer network parameters of 0.35M and less FLOPs of 18G, but yields $67.1\%$ mIoU that is $8.8\%$ higher than ENet \cite{Paszke2016ENet} without using any extra data and testing augmentation. The visualized segmentation results are presented in Figure \ref{Figure compare}. It can be seen that compared with ENet \cite{Paszke2016ENet}, the edges of objects segmented by EADNet are clearer. This shows that our network has stronger ability to segment the edges of objects.

\subsection{Comparison on Other Dataset}

We also evaluate our EADNet on CamVid dataset. We adopt the same setting as Cityscapes dataset, and the results are illustrated in Table \ref{Table 3}, which consistently show that our proposed EADNet achieves higher mIoU with significantly less FLOPs than state-of-the-art DFANet \cite{Li_2019_CVPR}.

\section{CONCLUSIONS}
In this work, we propose a multi-scale multi-shape receptive field convolution block and further apply this block to develop a lightweight semantic segmentation network called EADNet, which is suitable to deploy on computing resource constrained embedded devices. As a result, the EADNet reaches competitive segmentation accuracy with fewer parameters and less computation FLOPs than state-of-the-art works. Quantitative and qualitative experimental results on Cityscapes well demonstrate the effectiveness of our proposed method.

\vfill\pagebreak

\bibliographystyle{IEEEbib}
\bibliography{strings,refs}

\end{document}